\documentclass[10pt,twocolumn,letterpaper]{article}

\usepackage[pagenumbers]{cvpr} 

\usepackage{graphicx}
\usepackage{amsmath}
\usepackage{amssymb}
\usepackage{booktabs}

\usepackage{caption}
\usepackage{array}
\usepackage{multirow}
\usepackage{threeparttable}
\usepackage[dvipsnames]{xcolor}

\newcommand*\rot{\rotatebox{90}}

\usepackage[pagebackref,breaklinks,colorlinks]{hyperref}

\usepackage[capitalize]{cleveref}
\crefname{section}{Sec.}{Secs.}
\Crefname{section}{Section}{Sections}
\Crefname{table}{Table}{Tables}
\crefname{table}{Tab.}{Tabs.}

\def\cvprPaperID{xxxx} 
\def\confName{CVPR}
\def\confYear{2023}

\def\ourapproach{SiNC }

\begin{document}

\title{Non-Contrastive Unsupervised Learning of Physiological Signals from Video}

\author{Jeremy Speth, Nathan Vance, Patrick Flynn, Adam Czajka\\
University of Notre Dame\\
{\tt\small \{jspeth,nvance1,flynn,aczajka\}@nd.edu}
}
\maketitle

\begin{abstract}
Subtle periodic signals such as blood volume pulse and respiration can be extracted from RGB video, enabling remote health monitoring at low cost.
Advancements in remote pulse estimation -- or remote photoplethysmography (rPPG) -- are currently driven by deep learning solutions.
However, modern approaches are trained and evaluated on benchmark datasets with associated ground truth from contact-PPG sensors.
We present the first non-contrastive unsupervised learning framework for signal regression to break free from the constraints of labelled video data.
With minimal assumptions of periodicity and finite bandwidth, our approach is capable of discovering the blood volume pulse directly from unlabelled videos.
We find that encouraging sparse power spectra within normal physiological bandlimits and variance over batches of power spectra is sufficient for learning visual features of periodic signals.
We perform the first experiments utilizing unlabelled video data not specifically created for rPPG to train robust pulse rate estimators.
Given the limited inductive biases and impressive empirical results, the approach is theoretically capable of discovering other periodic signals from video, enabling multiple physiological measurements without the need for ground truth signals. Codes to fully reproduce the experiments are made available along with the paper.
\end{abstract}
\section{Introduction}
Camera-based vitals estimation is a rapidly growing field enabling non-contact health monitoring in a variety of settings~\cite{McDuff2022}.
Although many of the signals avoid detection from the human eye, video data in the visible and infrared ranges contain subtle intensity changes caused by physiological oscillations such as blood volume and respiration.
Significant remote photoplethysmography (rPPG) research for estimating the cardiac pulse has leveraged supervised deep learning for robust signal extraction~\cite{Chen2018,Yu2019,Niu2020,Liu_MTTS_2020,Speth_CVIU_2021,Yu_2022_CVPR}.
While the number of successful approaches has rapidly increased, the size of benchmark video datasets with simultaneous vitals recordings has remained relatively stagnant.

Robust deep learning-based systems for deployment require training on larger volumes of video data with diverse skin tones, lighting, camera sensors, and movement.
However, collecting simultaneous video and physiological ground truth with contact-PPG or electrocardiograms (ECG) is challenging for several reasons.
First, many hours of high quality videos is an unwieldy volume of data. Second, recording a diverse subject population in conditions representative of real-world activities is difficult to conduct in the lab setting. Finally, synchronizing contact measurements with video is technically challenging, and even contact measurements used for ground truth contain noise.

\definecolor{myorange}{RGB}{220, 112, 3}
\definecolor{mygreen}{RGB}{47, 154, 47}
\definecolor{myblue}{RGB}{115, 60, 160}

\begin{figure*}
    \centering
    \includegraphics[width=\linewidth]{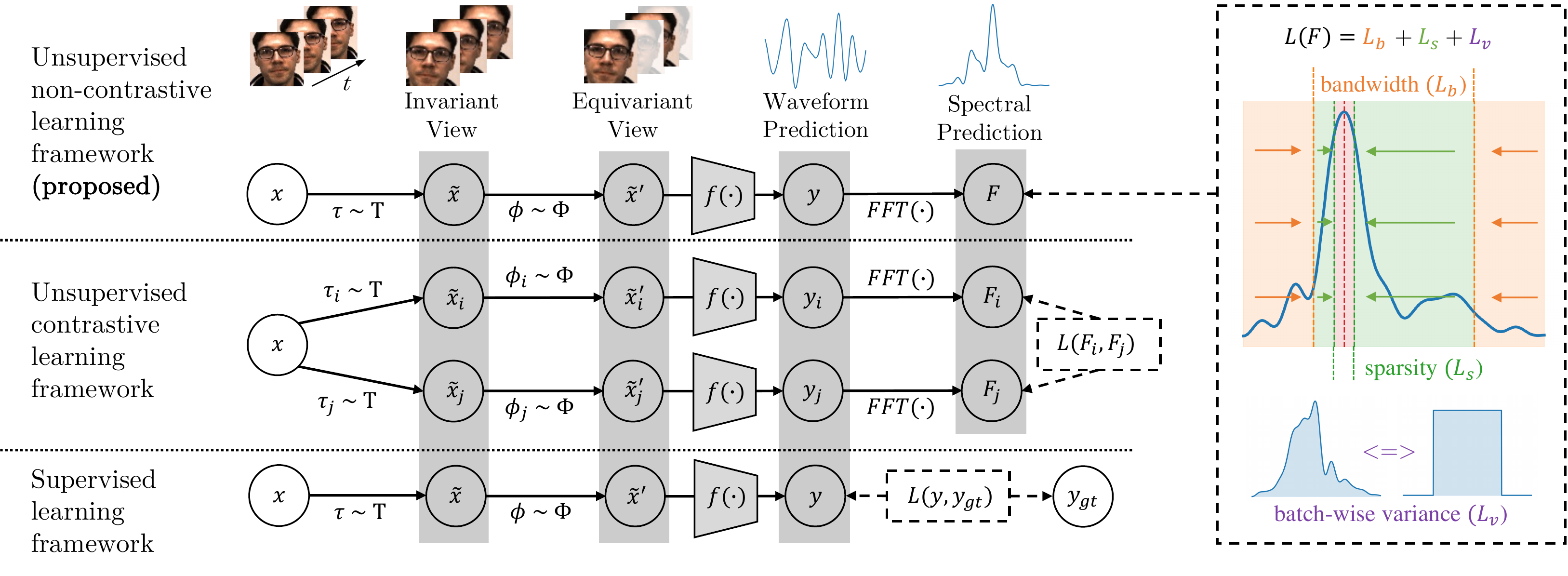}
    \caption{Overview of the \ourapproach framework for rPPG compared with traditional supervised and unsupervised learning. Supervised and contrastive losses use distance metrics to the ground truth or other samples. Our framework applies the loss directly to the prediction by shaping the frequency spectrum, and encouraging \textcolor{myblue}{{\bf variance over a batch}} of inputs. Power outside of the \textcolor{myorange}{{\bf bandlimits}} is penalized to learn invariances to irrelevant frequencies. Power within the bandlimits is encouraged to be \textcolor{mygreen}{{\bf sparsely}} distributed near the peak frequency.}
    \label{fig:framework}
\end{figure*}

Fortunately, recent works find that contrastive unsupervised learning for rPPG is a promising solution to the data scarcity problem~\cite{Gideon_2021_ICCV,Sun_2022_ECCV,Wang_SSL_2022,Yuzhe_SimPer_2022}.
We extend this research line into {\it non-contrastive} unsupervised learning to discover periodic signals in video data.
With end-to-end unsupervised learning, collecting more representative training data to learn powerful visual features is much simpler, since only video is required without associated medical information.

In this work, we show that non-contrastive unsupervised learning is especially simple when regressing rPPG signals.
We find weak assumptions of periodicity are sufficient for learning the minuscule visual features corresponding to the blood volume pulse from unlabelled face videos.
The loss functions can be computed in the frequency domain over batches without the need for pairwise or triplet comparisons.
Figure \ref{fig:framework} compares the proposed approach with supervised and contrastive unsupervised learning approaches.

This work creates opportunities for scaling deep learning models for camera-based vitals and estimating periodic or quasi-periodic signals from unlabelled data beyond rPPG.
Our {\bf novel contributions} are:
\begin{enumerate}
    \item A general framework for physiological {\bf si}gnal estimation via {\bf n}on-{\bf c}ontrastive unsupervised learning (SiNC) by leveraging periodic signal priors.
    \item The first {\bf non-contrastive} unsupervised learning method for camera-based vitals measurement.
    \item The first experiments and results of training with {\bf non-rPPG-specific video data} without ground truth vitals.
\end{enumerate}

Source code to replicate this work is available at \url{https://github.com/CVRL/SiNC-rPPG}.
\section{Related Work}

\subsection{Remote Photoplethysmography (rPPG)}
The primary class of approaches for remote pulse estimation have shifted over the last decade from blind source separation~\cite{Poh2010,Poh2011}, through linear color transformations~\cite{DeHaan2013,DeHaan2014,Wang2017,Wang2019} to training supervised deep learning-based models~\cite{Niu2018,Chen2018,Yu2019,Niu2020,Liu_MTTS_2020,Lee_ECCV_2020,Lu2021,Speth_CVIU_2021,Zhao_2021,Yu_2022_CVPR}. While the color transformations generalize well across many datasets, deep learning-based models give better accuracy when tested on data from a similar distribution to the training set. To this end, deep learning research has focused on optimizing neural architectures for extracting robust spatial and temporal features from the limited benchmark datasets.

To get around the data bottleneck, large synthetic physiological datasets have recently been proposed~\cite{Kadambi2022,mcduff2022scamps}. The SCAMPS dataset~\cite{mcduff2022scamps} contains videos for 2,800 synthetic avatars in various environments with a range of corresponding labels including PPG, EKG, respiration, and facial action units. The UCLA-synthetic dataset~\cite{Kadambi2022} contains 480 videos, and they show that training models with real and synthetic data gives the best results. Another strength of synthetic datasets is their ability to cover the broad range of skin tones, which may be difficult when collecting real data.

Another solution to the lack of physiological data is unsupervised learning, where a large set of videos and periodic priors on the output signal is sufficient~\cite{Gideon_2021_ICCV,Sun_2022_ECCV,Wang_SSL_2022,Yuzhe_SimPer_2022}. We discuss these methods in more detail in section \ref{sec:background_SSL_rPPG}.

\subsection{Unsupervised Learning}
Self-supervised learning is progressing quickly for image representation learning. Two main classes of approaches have been competing: contrastive and non-contrastive (or regularized) learning~\cite{bardes2022vicreg}. Contrastive approaches~\cite{Misra2019,Caron2020, Chen2020} define criteria for distinguishing whether two or more samples are the same or different, then pull or push the predicted embeddings. Non-contrastive methods augment positive pairs, and enforce variance in the predictions over batches to avoid {\it collapse}, in which the model's embeddings reside in a small subspace of the feature space, instead of spanning a larger or entire embedding space~\cite{bardes2022vicreg}. Distillation methods only use positive samples and avoid collapse by applying a moving average and stop-gradient operator~\cite{Grill2020,Chen2021}. Another class of approaches maximize information content of embeddings~\cite{Zbontar2021,ermolov2021whitening}.

\subsection{Unsupervised Learning for rPPG}\label{sec:background_SSL_rPPG}
All existing unsupervised rPPG approaches are contrastive~\cite{Gideon_2021_ICCV,Sun_2022_ECCV,Wang_SSL_2022,Yuzhe_SimPer_2022}. In the contrastive framework, pairs of input videos are input to the same model, and the predictions over similar videos are pulled closer, while the predictions from dissimilar videos are repelled.

Gideon \etal~\cite{Gideon_2021_ICCV} were the first to train a deep learning model without labels for rPPG using the contrastive framework. The core of their approach is frequency resampling to create negative samples. Although spatially similar, a resampled video clip contains a different underlying pulse rate, so the model must learn to attend to the temporal dynamics.
For their distance function between pairs, they calculated the mean square error between power spectral densities of the model's waveform predictions. While their approach learns to estimate the pulse, their formulation with negative samples is imprecise.
The resampled frequency for the negative sample is known, so the relative translation of the power spectrum from the anchor sample can be directly computed. Thus, rather than repelling the estimated spectra, it is more accurate to penalize differences from the known spectra. Furthermore, resampling close to the original sampling rate causes overlap in the power spectra, so repelling the pair is inaccurate.

Yuzhe \etal~\cite{Yuzhe_SimPer_2022} incorporated the previously ignored resampling factor for a soft modification to the InfoNCE loss~\cite{Oord2018} that scales the desired similarity between pairs by their relative sampling rate. A downside is that their learning framework is not end-to-end unsupervised and requires fine-tuning with PPG labels after the self-supervised stage.

Differently from \cite{Gideon_2021_ICCV}, Contrast-Phys~\cite{Sun_2022_ECCV} and SLF-RPM~\cite{Wang_SSL_2022} consider all samples different from the anchor to be negatives. This assumes that the power spectra will vary between subjects or sufficiently long windows for the same subject.
This runs into similar issues with negative pairs as Gideon's approach. Different subjects may have the same pulse rate, so punishing the model for predicting similar frequencies is common during training.
Furthermore, the Fast Fourier Transform (FFT) does not produce perfectly sparse decompositions, resulting in spectral overlap even if the heart rate differs by several beats per minute (bpm). As an example, the last column of the second row in Fig. \ref{fig:losses} shows the nulls of the main lobe are nearly 30 bpm apart.

\section{Method}
We first formulate the general setup for signal regression from video. A video sample $x_i \in \mathbb{R}^{T\times W\times H\times C}$ sampled from a dataset $\mathcal{D}$ consists of $T$ images of size $W\times H$ pixels across $C$ channels, captured uniformly over time.
State-of-the-art methods offer models $f$ that regress a waveform $\mathbb{R}^T \ni y_i = f(x_i)$ of the same length as the video.
Recently, the task has been effectively modeled end-to-end with the models $f$ being spatiotemporal neural networks~\cite{Yu2019,Lee_ECCV_2020,Speth_CVIU_2021,Lu_2021,Yu_2022_CVPR}. While most previous works are supervised and minimize the loss to a contact pulse measurement, we perform non-contrastive learning using only the model's estimated waveform.

The key realization is that we can place strong priors on the estimated pulse regarding its bandwidth and periodicity.
Observed signals outside the desired frequency range are pollutants, so penalizing the model for carrying them through the forward pass results in invariances to such noisy visual features.
We find that it is surprisingly easy to impose the desired constraints in the frequency domain.
Thus, all waveforms are transformed into their discrete Fourier components with the FFT before computing all losses in our approach, $F = \mathrm{FFT}(y)$. The following sections describe the loss functions and augmentations used during training.

\subsection{Losses}

\begin{figure*}
    \centering
    \includegraphics[width=\linewidth]{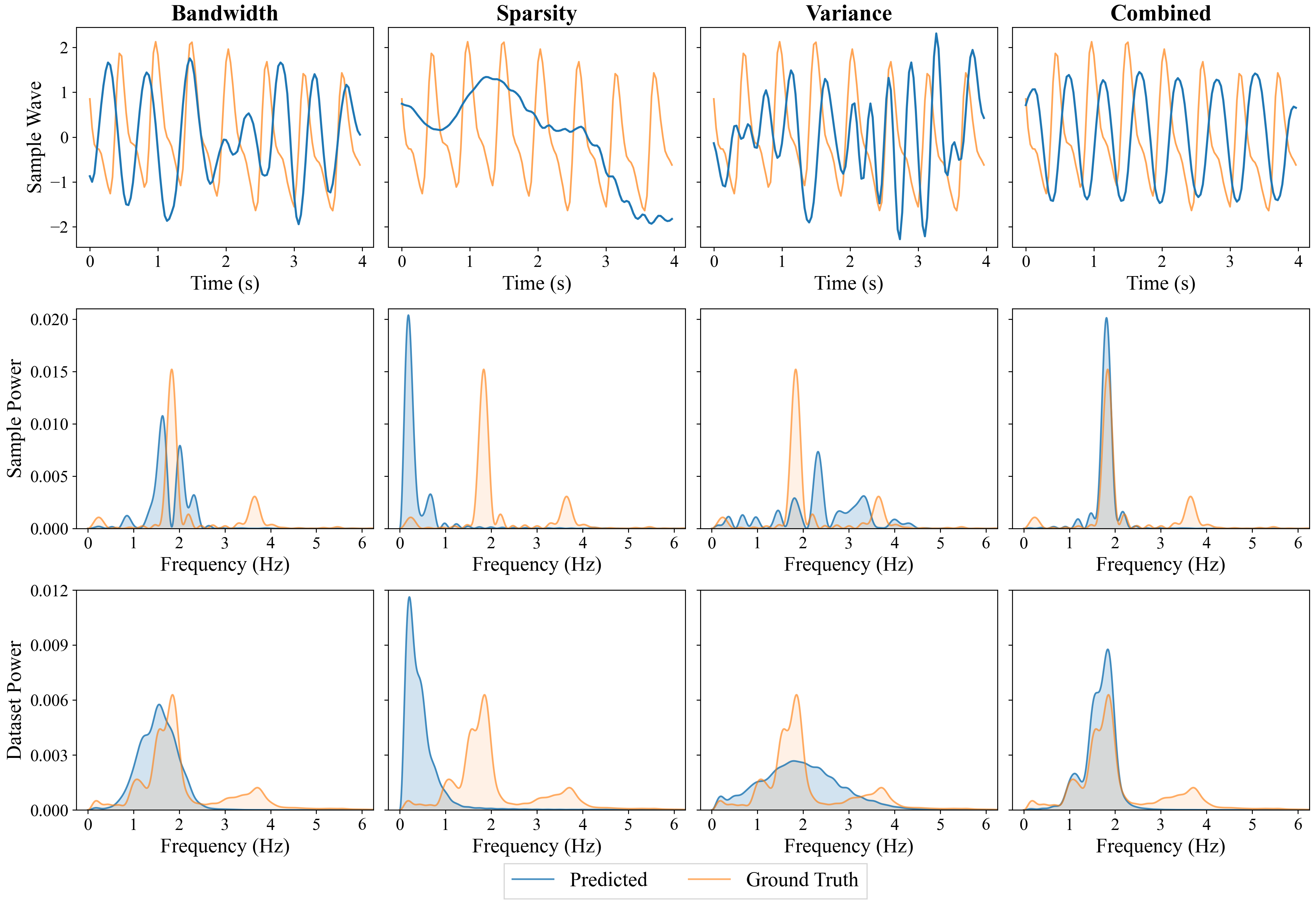}
    \caption{Each column shows predictions from models trained with one or all of the losses for 20 epochs on UBFC-rPPG. The first two rows show a sample in the time and frequency domain, respectively. The last row shows the signal power over the validation set computed by taking the sum of normalized power spectral densities from each sample, then dividing the result by the number of validation samples. The \textbf{bandwidth} loss penalizes signal power outside predefined bandlimits (40 to 180 bpm) to constrain the output space. The \textbf{sparsity} loss encourages a narrow spectrum containing a strong periodic component. By itself, the model learns the simplest solution characterized by very low frequencies. The \textbf{variance} loss encourages diverse power spectra over a batch, preventing the model from collapsing to a narrow bandwidth. When \textbf{combined}, the model learns to estimate periodic signals within the desired bandlimits.}
    \label{fig:losses}
\end{figure*}

One of the advantages of unsupervised learning for periodic signals is that we can constrain the solution space significantly. For physiological signals such as respiration and blood volume pulse, we know the healthy upper and lower bounds of the frequencies. We also desire the extracted signal to be sparse in the frequency domain, and that our model filters out noise signals present in the video. With these constraints, we can greatly simplify the problem of finding good features for the desired signal in the data.

\subsubsection{Bandwidth Loss}
One of the most powerful constraints we can place on the model is frequency bandlimits. Past unsupervised methods have used the irrelevant power ratio (IPR) as a validation metric~\cite{Gideon_2021_ICCV,Gideon_2021_ICCVW,Sun_2022_ECCV} for model selection. We find that it is also effective during model training. The IPR penalizes the model for generating signals outside the desired bandlimits. With lower and upper bandlimits of $a$ and $b$, respectively, our bandwidth loss becomes:
\begin{equation}\label{eq:L_b}
    L_b = \dfrac{\sum\limits_{i=-\infty}^a F_i + \sum\limits_{i=b}^{\infty} F_i}{\sum\limits_{i=-\infty}^\infty F_i},
\end{equation}
\noindent
where $F_i$ is the power in the $i$th frequency bin of the predicted signal. This simple loss enforces learning of many invariants, such as movement from respiration, talking, or facial expressions which typically occupy low frequencies. In our experiments we specify the limits as $a=0.\overline{66}$ Hz to $b=3$ Hz, which corresponds to a common pulse rate range from 40 bpm to 180 bpm. The first column of Fig. \ref{fig:losses} shows the result of training exclusively with the bandwidth loss $L_b$. The last row shows that the model concentrates signal power between the bandlimits.

\subsubsection{Sparsity Loss}
The pulse rate is the most common physiological marker associated with the blood volume pulse. Since we are primarily interested in the frequency, we can further improve our model by preventing wideband predictions. This also reveals the true signal we aim to discover by ignoring visual dynamics that are not strongly periodic.

We penalize energy within the bandlimits that is not near the spectral peak:
\begin{equation}\label{eq:L_s}
    L_s = \dfrac{\sum\limits_{i=a}^{\mathrm{argmax}(F) - \Delta_F} F_i + \sum\limits_{i=\mathrm{argmax}(F) + \Delta_F}^b F_i}{\sum\limits_{i=a}^b F_i},
\end{equation}
where $\mathrm{argmax}(F)$ is the frequency of the spectral peak, and $\Delta_F$ is the frequency padding around the peak.
All experiments are performed with a $\Delta_F$ of 6 beats per minute~\cite{Nowara_BOE_2021}. Figure \ref{fig:losses} shows result of training only with the sparsity loss in the second column. For a single sample, the power spectrum is very sparse. For the whole dataset, we see that it is easiest to predict sparse solutions when high frequencies are ignored entirely and visual features corresponding to low frequencies are learned. 

\subsubsection{Variance Loss}
One of the risks of non-contrastive methods is the model collapsing into trivial solutions and making predictions independently of the input features. In regularized methods such as VICReg~\cite{bardes2022vicreg}, a hinge loss on the variance over a batch of predictions is used to enforce diverse outputs. We use a similar strategy to avoid model collapse, but instead spread the variance in power spectral densities towards a uniform distribution over the desired frequency band.
Our variance loss processes a uniform prior distribution $P$ over $d$ frequencies, and a batch of $n$ spectral densities, $F = [v_1,...,v_n]$, where each vector is a $d$-dimensional frequency decomposition of a predicted waveform. We calculate the normalized sum of densities over the batch, $Q$, and define the variance loss as the squared Wasserstein distance~\cite{hou_emd_2017} to the uniform prior:
\begin{equation}
    L_v = \frac{1}{d} \sum_{i=1}^d \left( \mathrm{CDF}_i(Q) - \mathrm{CDF}_i(P) \right)^2,
\end{equation}
\noindent
where CDF is a cumulative distribution function. The third column of Fig. \ref{fig:losses} shows the effect of the variance loss during training. For a single sample, wide-band signals containing multiple frequencies are predicted, and the frequencies over the predicted dataset cover the task's bandwidth. In our experiments we use a batch size of 20 samples. See Appendix \ref{sec:app_batch_size} for an ablation experiment on the impact of smaller batch sizes.

\subsubsection{Combining All Losses}
Summarizing, our training loss function is a simple sum of the aforementioned losses:
\begin{equation}
    L = L_b + L_s + L_v.
\end{equation}
While one could weight particular components of the loss more than others, we specifically formulated the losses to scale them between 0 and 1. In our experiments, we find that a simple summation without weighting gives good performance. \textbf{The combined loss function encourages the model to search over the supported frequencies to discover visual features for a strong periodic signal.} Remarkably, we find that this simple framework is sufficient for learning to regress the blood volume in video, as shown in the last column of Fig. \ref{fig:losses}.

\subsection{Augmentations}\label{sec:augmentations}
Unlike Gideon \etal's~\cite{Gideon_2021_ICCV} approach, which only applies frequency augmentations, we apply several augmentations to both the spatial and temporal dimensions to learn invariances to noisy visual signals. In fact, we found that without augmentations, models did not converge during training (see Appendix \ref{sec:app_no_augmentations}).

{\bf Image Intensity Augmentations.} Random Gaussian noise is added to each pixel location in a clip with a mean of 0 and a standard deviation of 2 on the original image scale from 0 to 255. The illumination is augmented by adding a constant sampled from a Gaussian distribution with mean 0 and standard deviation of 10 to every pixel in a clip, which darkens or brightens the video.

{\bf Spatial Augmentations.} We randomly horizontally flip a video clip with 50\% probability. The spatial dimension of a clip are randomly square cropped down to between half the original length and the original length. The cropped clip is then linearly interpolated back to the original dimensions.

{\bf Temporal Augmentations.} With the general assumption that the desired signal is strongly periodic and sparsely represented in the Fourier domain, we randomly flip a video clip along the time dimension with a probability of 50\%. Note that the Fourier decomposition of a time-reversed sinusoid is identical to that of the original sinusoid.

{\bf Frequency Augmentations.} Perhaps the most important augmentation is frequency resampling~\cite{Gideon_2021_ICCV}, where the video is linearly interpolated to a different frame rate. This augmentation is particularly interesting for rPPG, because it transforms the video input and target signal equivalently along the time dimension, making it equivariant. Given the aforementioned transformations that are invariant, $\tau(\cdot) \sim \mathcal{T}$, the equivariant frequency resampling operation, $\phi(\cdot) \sim \Phi$, and a model $f(\cdot)$ that infers a waveform from a video we have the following:
\begin{equation}
    \phi(f(\tau(x))) = f(\phi(\tau(x))).
\end{equation}

This is a powerful augmentation, because it allows us to augment the target distribution along with the video input. In our experiments we randomly resample input clips by a factor $c \sim U(0.6, 1.4)$. After applying the resampling augmentation, we scale the bandlimits by $c$, to avoid penalizing the model if the augmentation pushed the underlying pulse frequency outside of the original bandlimits.



\section{Datasets}
We use PURE~\cite{Stricker2014}, UBFC-rPPG~\cite{Bobbia2019}, and DDPM~\cite{Speth_IJCB_2021} as benchmark rPPG datasets for training and testing, and CelebV-HQ dataset~\cite{zhu2022celebvhq} and HKBU-MARs~\cite{Liu_2016_CVPRW} for unsupervised training only.

\noindent
\textbf{ Deception Detection and Physiological Monitoring (DDPM)}~\cite{Speth_IJCB_2021,Vance2022} consists of 86 subjects in an interview setting, where subjects attempted to answer questions deceptively. Interviews were recorded at 90 frames-per-second for more than 10 minutes on average. Natural conversation and frequent head pose changes make it a difficult and less-constrained rPPG dataset.

\textbf{PURE}~\cite{Stricker2014} is a benchmark rPPG dataset consisting of 10 subjects recorded over 6 sessions. Each session lasted approximately 1 minute, and raw video was recorded at 30 fps. The 6 sessions for each subject consisted of: (1) steady, (2) talking, (3) slow head translation, (4) fast head translation, (5) small and (6) medium head rotations. Pulse rates are at or close to the subject's resting rate.

\textbf{UBFC-rPPG}~\cite{Bobbia2019} contains 1-minute long videos from 42 subjects recorded at 30 fps. Subjects played a time-sensitive mathematical game to raise their heart rates, but head motion is limited during the recording.

\textbf{HKBU 3D Mask Attack with Real World Variations (HKBU-MARs)}~\cite{Liu_2016_CVPRW} consists of 12 subjects captured over 6 different lighting configurations with 7 different cameras each, resulting in 504 videos lasting 10 seconds each. The diverse lighting and camera sensors make it a valuable dataset for unsupervised training. We use version 2 of HKBU-MARs, which contains videos with both realistic 3D masks and unmasked subjects.

\textbf{High-Quality Celebrity Video Dataset (CelebV-HQ)}~\cite{zhu2022celebvhq} is a set of processed YouTube videos containing 35,666 face videos from over 15,000 identities. The videos vary dramatically in length, lighting, emotion, motion, skin tones, and camera sensors. The greatest challenge in harnessing online videos is their reduced quality due to compression before upload and by the video provider. Compression is a known challenge for rPPG, since the blood volume pulse is so subtle optically~\cite{McDuff2017,Nowara_ICCVW_2019,Rapczynski2019,Nowara_BOE_2021}.
\section{Training Details}

\setlength\tabcolsep{4pt}
\begin{table*}[htb!]
\caption{{\bf Intra-dataset} pulse rate estimation results. The best results for a given metric/dataset are bolded, and the second-best results are underlined. For a better comparison with results in the literature, this table follows \cite{Sun_2022_ECCV}. MAE: Mean Absolute Error; RMSE: Root Mean Square Error; $r$: Pearson correlation coefficient.}
\fontsize{7.533}{9}\selectfont
\centering
{
\begin{threeparttable}
\begin{tabular}{llccccccccc}
\toprule
\multirow{2}{*}{\begin{tabular}[c]{@{}l@{}}\\ Types\end{tabular}} &
\multirow{2}{*}{\begin{tabular}[c]{@{}l@{}}\\Methods\end{tabular}} &
\multicolumn{3}{c}{UBFC-rPPG} & \multicolumn{3}{c}{PURE} &
\multicolumn{3}{c}{DDPM} \\ 
\cmidrule(lr){3-5}\cmidrule(lr){6-8}\cmidrule(lr){9-11}
& & \begin{tabular}[c]{@{}c@{}}MAE\\ (bpm)\end{tabular} & \begin{tabular}[c]{@{}c@{}}RMSE\\ (bpm)\end{tabular} & $r$                    & \begin{tabular}[c]{@{}c@{}}MAE\\ (bpm)\end{tabular} & \begin{tabular}[c]{@{}c@{}}RMSE\\ (bpm)\end{tabular} & $r$                    & \begin{tabular}[c]{@{}c@{}}MAE\\ (bpm)\end{tabular} & \begin{tabular}[c]{@{}c@{}}RMSE\\ (bpm)\end{tabular} & $r$                  \\ 
\midrule
\multirow{5}{*}{\rot{Traditional}}
& GREEN \cite{Verkruysse2008} & 7.50 & 14.41 & 0.62 & 7.23 & 17.05 & 0.69 & 32.79 & 43.09 & 0.04 \\
& ICA \cite{Poh2011} & 5.17 & 11.76 & 0.65 & 3.76 & 12.60 & 0.85 & 22.22 & 35.77 & 0.40 \\
& CHROM \cite{DeHaan2013} &
    2.36 & 9.23 & 0.87 &
    \underline{0.75} & 2.23 & \bf 1.00 &
    13.48 & 28.53 & 0.56 \\
& 2SR \cite{Wang2016} & - & - & - &
    2.44 & 3.06 & 0.98 & - & - & - \\
& POS \cite{Wang2017} &
    2.11 & 9.11 & 0.87 &
    0.80 & 4.11 & 0.98 &
    9.03 & 23.07 & 0.70 \\
 
\midrule
\multirow{6}{*}{\rot{Supervised}}

& HR-CNN \cite{Spetlik_2018} & - & - & - & 1.84 & 2.37 & 0.98 & - & - & - \\
& SynRhythm \cite{Niu2018} & 5.59 & 6.82 & 0.72 & - & - & - & - & - & - \\
& PulseGAN \cite{Song_2021} & 1.19 & 2.10 & \underline{0.98} & - & - & - & - & - & - \\
& Dual-GAN \cite{Lu_2021} &
    \bf{0.44} & \bf{0.67} & \bf{0.99} &
    0.82 & \bf{1.31} & \underline{0.99} &
    - & - & - \\
& RPNet \cite{Speth_CVIU_2021}\tnote{\textdagger} &
    \underline{0.53 $\pm$ 0.01} & 1.78 $\pm$ 0.02 & \bf{0.99 $\pm$ 0.00} &
    1.15 $\pm$ 0.27 & 5.77 $\pm$ 1.25 & 0.96 $\pm$ 0.02 &
    \bf 3.46 $\pm$ 0.24 & \bf 12.47 $\pm$ 0.68 & \bf 0.91 $\pm$ 0.01 \\
& PhysNet \cite{Yu2019}\tnote{\textdagger} &
    0.55 $\pm$ 0.03 & 2.03 $\pm$ 0.37 & \bf 0.99 $\pm$ 0.00 &
    0.99 $\pm$ 0.19 & 5.22 $\pm$ 0.93 & 0.97 $\pm$ 0.01 &
    \underline{3.96 $\pm$ 0.76} & \underline{13.57 $\pm$ 1.74} & 
    \underline{0.89 $\pm$ 0.03} \\
\midrule

\multirow{5}{*}{\rot{Unsupervised}}
 & Gideon2021 \cite{Gideon_2021_ICCV} & 
    1.85 & 4.28 & 0.93 &
    2.3 & 2.9 & \underline{0.99} &
    - & - & - \\
 & SLF-RPM~\cite{Wang_SSL_2022}\tnote{*} &
    8.39 & 9.70 & 0.70 &
    - & - & - &
    - & - & - \\
 & SimPer~\cite{Yuzhe_SimPer_2022}\tnote{*} &
    4.24 & - & - &
    3.89 & - & - &
    - & - & - \\
 & Contrast-Phys~\cite{Sun_2022_ECCV} &
    0.64 & \underline{1.00} & \bf{0.99} &
    1.00 & \underline{1.40} & \underline{0.99} &
    9.70 $\pm$ 2.90 & 25.02 $\pm$ 6.01 & 0.58 $\pm$ 0.19 \\
 & \bf{\ourapproach (ours)} &
    0.59 $\pm$ 0.00 & 1.83 $\pm$ 0.04 & \bf 0.99 $\pm$ 0.00 &
    \bf 0.61 $\pm$ 0.06 & 1.84 $\pm$ 0.40 & \bf 1.00 $\pm$ 0.00 &
    5.87 $\pm$ 0.11 & 17.44 $\pm$ 0.16 & 0.81 $\pm$ 0.00 \\
\bottomrule
\end{tabular}
\begin{tablenotes}\footnotesize
        \item[*] Some methods in the unsupervised row require fine-tuning on labeled data with a linear classifier.
        \item[\textdagger] Some supervised methods were trained with identical data augmentations to \ourapproach for fair comparison.
        \end{tablenotes}
    \end{threeparttable}
}
\label{tab:within_dataset}
\end{table*}

\subsection{Data Preprocessing}
To prepare the video clips for the spatiotemporal deep learning models, we first extract 68 face landmarks with OpenFace~\cite{Baltrusaitis2018}. We then define a bounding box in each frame with the minimum and maximum $(x,y)$ locations by extending the crop horizontally by 5\% to ensure that the cheeks and jaw are present. The top and bottom are extended by 30\%  and 5\% of the bounding box height, respectively, to include the forehead and jaw. We further extend the shorter of the two axes to the length of the other to form a square. The cropped frames are then resized to 64$\times$64 pixels with bicubic interpolation. For faster processing of the massive CelebV-HQ~\cite{zhu2022celebvhq} dataset, we instead use MediaPipe Face Mesh~\cite{lugaresi2019mediapipe} for landmarking.

\subsection{Model Architectures}
We use a 3D-CNN architecture similar to \cite{Speth_CVIU_2021} without temporal dilations, which was originally inspired by \cite{Yu2019}. We use a temporal kernel width of 5, and replace default zero-padding by repeating the edges. Zero-padding along the time dimension can result in edge effects that add artificial frequencies to the predictions. Early experiments showed that temporal dilations caused aliasing and reduced the bandwidth of the model to specific frequencies. Our losses and framework may be applied to any task and architecture with dense predictions along one or more dimensions. However, popular rPPG architectures such as DeepPhys~\cite{Chen2018} and  MTTS-CAN~\cite{Liu_MTTS_2020} may be ill-suited for the approach, since they consume very few frames, and the number of time points should be large enough to give sufficient frequency resolution with the FFT. In our experiments, we use the AdamW~\cite{Loshchilov2019AdamW} optimizer with a learning rate of 0.0001. We use a clip length of $T=120$ frames (4 seconds), and we set the input signal’s length to achieve a frequency resolution of $0.\overline{33}$ bpm.

\subsection{Supervised Training}
To properly compare our approach to its supervised counterpart we use the same model architecture and train it with the commonly used negative Pearson loss between the predicted waveform and the contact sensor ground truth~\cite{Yu2019}. During training we apply all of the same augmentations except time reversal. Models are trained for 200 epochs on PURE and UBFC-rPPG, and for 40 epochs on DDPM. The model from the epoch with the lowest loss on the validation set is selected for testing.

\subsection{Unsupervised Training}
Unsupervised models are trained for the same number of epochs as the supervised setting for both PURE and UBFC-rPPG, but we train for an additional 40 epochs on DDPM, since this dataset is considerably more difficult. We set the batch size to 20 samples during training. Contrary to previous unsupervised approaches~\cite{Gideon_2021_ICCV,Sun_2022_ECCV}, we leverage validation sets for model selection by selecting the model with the lowest combined bandwidth and sparsity losses. The creation of the dataset splits is described in the next section.

\subsection{Evaluation}
Pulse rates are computed as the highest spectral peak between 0.66 Hz and 3 Hz (equivalent to 40 bpm to 180 bpm) over a 10-second sliding window. The same procedure is applied to the ground truth waveforms for a reliable evaluation~\cite{Mironenko2020}. We apply common error metrics such as mean absolute error (MAE), root mean square error (RMSE), and Pearson correlation coefficient ($r$) between the pulse rates.

We perform 5-fold cross validation for both PURE and UBFC with the same folds as \cite{Gideon_2021_ICCV}, and use the predefined dataset splits from DDPM~\cite{Speth_CVIU_2021}. Differently from \cite{Gideon_2021_ICCV}, we use 3 of the folds for training, 1 for validation, and the remaining for testing rather than only training and testing partitions. We train 3 models with different initializations, resulting in 15 models trained on PURE and UBFC each, and 3 models trained on DDPM. We present the mean and standard deviation of the errors in the results.
\section{Results}
\subsection{Within-Dataset Testing}
Table \ref{tab:within_dataset} shows the results for models trained and tested on subject-disjoint partitions from the same datasets. For PURE and UBFC we achieve MAE lower than 1 bpm, performing better or on par with all traditional and supervised learning approaches. For PURE, our approach gives the lowest MAE and a Pearson $r$ of nearly 1. Performance drops on DDPM due to the overall difficulty of the dataset. \ourapproach outperforms contrastive approaches, only being surpassed by supervised deep learning models.

In comparison to other unsupervised methods, Contrast-Phys~\cite{Sun_2022_ECCV} gives the most competitive performance on all but DDPM. Note that our approach gives the lowest MAE on all datasets, but has higher RMSE. We believe this is due their use of harmonic removal as a post-processing step when estimating the pulse rate, which is not described in \cite{Sun_2022_ECCV}, but can be found in their publicly available code. 

\subsection{Cross-Dataset Testing}
\begin{table}[!htb]
\setlength\tabcolsep{6pt}
\centering\footnotesize
\caption{{\bf Cross-dataset} pulse rate estimation performance. The top 3 training datasets are common rPPG benchmarks, while HKBU was not designed for rPPG and has no pulse ground truth. Note that testing on UBFC after training on DDPM performs well, since their frequency distributions are similar, while PURE's pulse rates tend to be much lower.}
\begin{tabular}{cccrr}
\toprule
\begin{tabular}{@{}c@{}}\textbf{Training}\\\textbf{Dataset}\end{tabular} & \begin{tabular}{@{}c@{}}\textbf{Testing}\\\textbf{Dataset}\end{tabular} & {\bf Method} & \begin{tabular}{@{}c@{}}{\bf MAE\phantom{xxi}}\\\textbf{(bpm)\phantom{xxi}}\end{tabular} & {\bf $r$\phantom{xxxx}}
\\
\cmidrule{1-5}
\multirow{6}{*}{DDPM}
& UBFC & \ourapproach & 0.88 $\pm$ 0.25 & 0.98 $\pm$ 0.01\\
& UBFC & Contrast-Phys & 1.14 $\pm$ 0.38 & 0.96 $\pm$ 0.03\\
& UBFC & PhysNet & 1.11 $\pm$ 0.42 & 0.95 $\pm$ 0.02 \\
\cline{2-5}\\[-2ex]
& PURE & \ourapproach & 3.12 $\pm$ 1.07 & 0.88 $\pm$ 0.05 \\
& PURE & Contrast-Phys & 13.02 $\pm$ 6.12 & 0.19 $\pm$ 0.59  \\
& PURE & PhysNet & 1.46 $\pm$ 0.34 & 0.95 $\pm$ 0.02 \\
\midrule
 
\multirow{6}{*}{UBFC}
& DDPM & \ourapproach & 18.53 $\pm$ 0.36 & 0.38 $\pm$ 0.01 \\
& DDPM & Contrast-Phys & 22.93 $\pm$ 1.02 & 0.18 $\pm$ 0.04 \\
& DDPM & PhysNet & 18.58 $\pm$ 0.12 & 0.40 $\pm$ 0.00 \\
\cline{2-5}\\[-2ex]
& PURE & \ourapproach & 4.02 $\pm$ 0.06 & 0.86 $\pm$ 0.00 \\
& PURE & Contrast-Phys & 19.61 $\pm$ 2.01 & 0.33 $\pm$ 0.06 \\
& PURE & PhysNet & 3.81 $\pm$ 0.34 & 0.87 $\pm$ 0.02 \\
\midrule

\multirow{6}{*}{PURE}
& UBFC & \ourapproach & 6.64 $\pm$ 1.76 & 0.59 $\pm$ 0.10 \\
& UBFC & Contrast-Phys & 10.22 $\pm$ 0.38 & 0.45 $\pm$ 0.04 \\
& UBFC & PhysNet & 7.02 $\pm$ 3.35 & 0.60 $\pm$ 0.13 \\
\cline{2-5}\\[-2ex]
& DDPM & \ourapproach & 24.92 $\pm$ 0.65 & 0.20 $\pm$ 0.00 \\
& DDPM & Contrast-Phys & 29.63 $\pm$ 0.48 & 0.03 $\pm$ 0.02 \\
& DDPM & PhysNet & 28.03 $\pm$ 2.20 & 0.13 $\pm$ 0.05\\
\midrule
\midrule

\multirow{3}{*}{\begin{tabular}{@{}c@{}}HKBU\\(non-rPPG)\end{tabular}}

& UBFC & \ourapproach &  1.08 $\pm$ 0.03 & 0.95 $\pm$ 0.00 \\
& PURE      & \ourapproach &  2.43 $\pm$ 0.20 & 0.90 $\pm$ 0.02 \\
& DDPM      & \ourapproach & 20.34 $\pm$ 0.25 & 0.19 $\pm$ 0.02 \\
\bottomrule
\end{tabular}
\label{tab:cross_dataset}
\end{table}

We perform cross-dataset testing to analyze whether the approach is robust to changes to the lighting, camera sensor, pulse rate distribution, and motion. Table \ref{tab:cross_dataset} shows the results for \ourapproach and supervised training with the same architecture. We find that the performance is similar for the supervised and unsupervised approaches when transferring to different data sources. Training on PURE exclusively gives relatively poor results when transferring to UBFC-rPPG and DDPM, due to the low pulse rate variability within PURE samples and lack of movement. Training on DDPM gives the best results overall, since the dataset is the largest and captures larger subjects' movements compared to other datasets.

\subsection{Training with CelebV-HQ Videos}\label{sec:celebv_training}
Given the abundance of face videos publicly available online, we trained a model on the CelebV-HQ dataset~\cite{zhu2022celebvhq}. After processing the available videos with MediaPipe and resampling to 30 fps, our unlabeled dataset consisted of 34,029 videos. We trained the model for 23 epochs and manually stopped training due to a plateau in the validation loss.
Unfortunately, we found that the model could not converge to the true blood volume pulse. We attribute the failure to poor video quality from compression. Although the videos were downloaded with the highest available quality, they have likely been compressed multiple times, removing the pulse signal entirely. See Appendix \ref{sec:app_celebv} for a quantitative assessment of rPPG quality using a baseline approach.

\subsection{Training with HKBU-MARs Videos}
The HKBU-MARs dataset~\cite{Liu_2016_CVPRW} was designed for face presentation attack detection, but we trained models on the ``real'' video sessions in the dataset. The bottom rows in Table \ref{tab:cross_dataset} show the results for training on HKBU-MARs, then testing on the benchmark rPPG datasets. Training on HKBU-MARs gives better results when testing on UBFC-rPPG and PURE than all training sets except DDPM, which is an order of magnitude larger. {\bf To our knowledge, this is the first succesful experiment showing that non-rPPG videos can be used to train robust rPPG models, even if they do not have ground-truth pulse labels}.

\begin{figure}
    \centering
    \includegraphics[width=\linewidth]{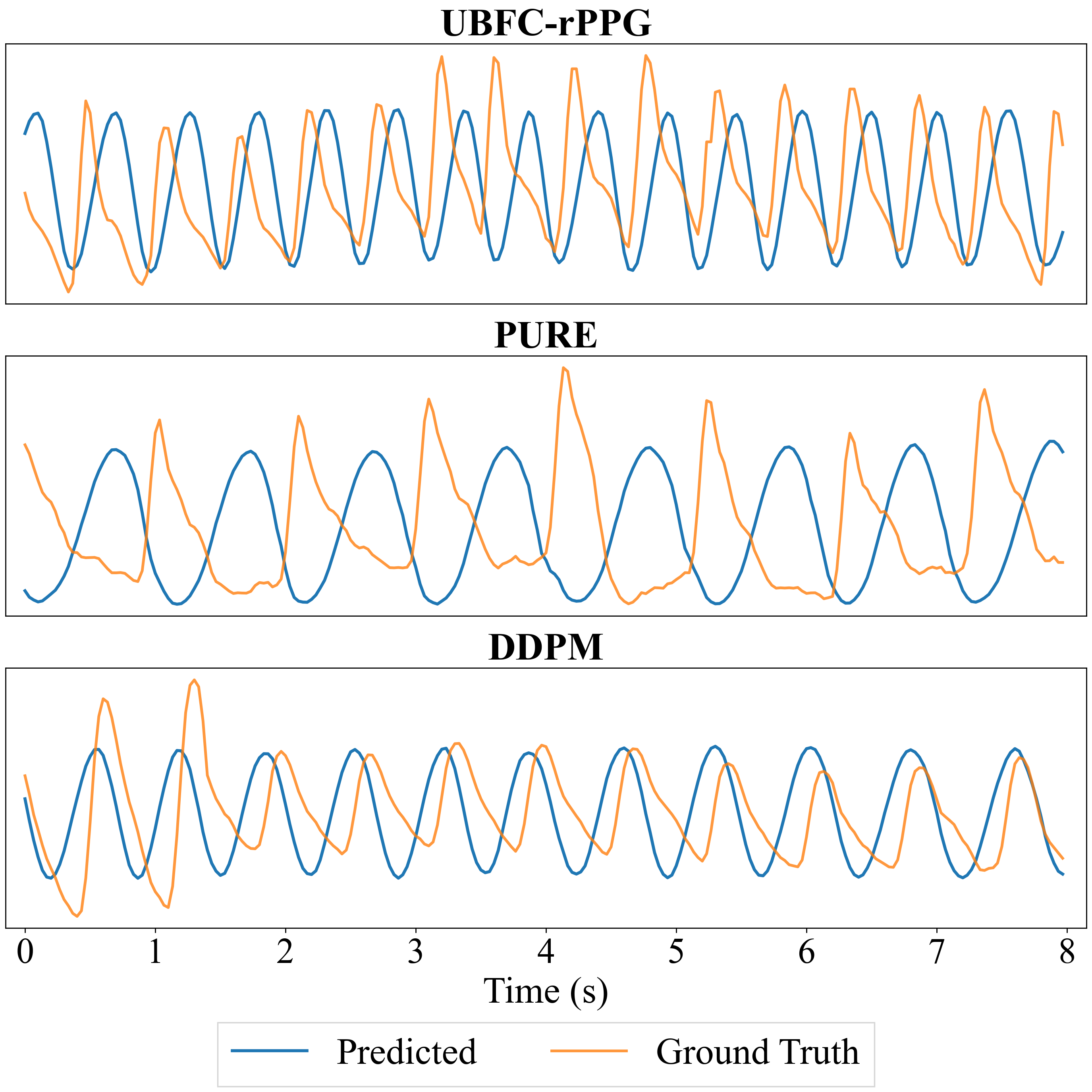}
    \caption{Within-dataset waveform predictions on all baseline datasets from end-to-end unsupervised models over an 8-second window. The model predictions are remarkably periodic without any form of filtering. Note that phase is not considered during training, so each model learns its own phase shift.}
    \label{fig:preds}
\end{figure}

\subsection{Ablation Study on Losses}
\begin{table}[!htb]
\setlength\tabcolsep{6pt}
\centering\footnotesize
\caption{Ablation study on the loss functions used during training. Results are shown for models trained and tested on UBFC-rPPG.}
\begin{tabular}{lrrr}
\toprule
{\bf Loss} & {\bf MAE (bpm)} & {\bf RMSE (bpm)} & {\bf $r$\phantom{xxxx}}
\\
\midrule
    $L_b$   &  3.08 $\pm$ 1.69 &  8.08 $\pm$ 3.61 &  0.87 $\pm 0.08$ \\
    $L_s$   & 45.50 $\pm$ 1.22 & 50.04 $\pm$ 0.94 & -0.04 $\pm 0.08$ \\
    $L_v$   & 22.89 $\pm$ 2.83 & 31.51 $\pm$ 2.36 &  0.22 $\pm 0.09$ \\
    $L_s+L_v$   & 51.24 $\pm$ 5.36 & 57.80 $\pm$ 7.39 & -0.04 $\pm 0.09$ \\
    $L_b+L_s$   &  9.99 $\pm$ 2.55 & 17.14 $\pm$ 2.36 &  0.51 $\pm 0.14$ \\
    $L_b+L_v$   &  4.18 $\pm$ 2.88 &  8.90 $\pm$ 5.24 &  0.82 $\pm 0.14$ \\
    \midrule
    \textbf{$L_b+L_s+L_v$}  & \bf 0.59 $\pm$ 0.00 & \bf 1.83 $\pm$ 0.04 & \bf 0.99 $\pm$ 0.00 \\
\bottomrule
\end{tabular}
\label{tab:loss_ablation}
\end{table}
We trained models using all combinations of loss components to analyze their contributions. Table \ref{tab:loss_ablation} shows the results for training and testing on UBFC-rPPG. The bandwidth loss is the most critical for discovering the true blood volume pulse, while the sparsity and variance losses do not learn the desired signal by themselves. Surprisingly, combining the bandwidth loss with just one of the sparsity or variance losses gives worse performance than just the bandwidth loss. However, when combining all three components, the model achieves impressive results.
\section{Discussion}

\subsection{Improvements over Supervised Learning}
It is initially surprising that unsupervised training leads to similar or improved rPPG estimation models compared to those trained in a supervised manner. However, there are several potential benefits to unsupervised training. From a hardware perspective, one of the difficulties in supervised training is aligning the contact pulse waveform with the video frames~\cite{Zhan2020}. The pulse sensor and camera may have a time lag, effectively giving the model an out-of-phase target at training time. Unsupervised training gives the model freedom to learn the phase directly from the video. The contact-PPG signal is also sensitive to motion and may be noisy. Since motion may co-occur at the face and fingertip, the contact signal may misguide the model towards visual features for which they should be invariant.

From a physiological perspective, the pulse observed optically at the fingertip with a contact sensor has a different phase than that of the face, since blood propagates along a different path before reaching the peripheral microvasculature, making alignment nearly impossible without shifting the targets to rPPG estimates from existing methods~\cite{Speth_CVIU_2021}. Additionally, the morphological shape of the contact-PPG waveform depends on numerous factors such as the wavelength of light (and corresponding tissue penetration depth), external pressure from the oximeter clip, and vasodilation at the measurement site~\cite{Moco2018,Abraham2013}. This indicates that the morphology and phase of the target PPG waveform is likely different from the observed rPPG waveform.

\subsection{Why Does It Work?}
The success of the proposed non-contrastive approach depends on specific properties of the data, model, and how the two interact.
Limited model capacity is actually a strength, since it forces discovering features to generalize across inputs.
An infinite capacity network could discover spurious signals in the training data and fail to generalize.
By constraining the model's predictions to have specific periodic properties the limited-capacity model must find a general set of features to produce a signal that exists in all of the training samples, which happens to be the blood volume pulse in our datasets.

As a beneficial side-effect, the model intrinsically learns to ignore common noise factors such as illumination, rigid motion, non-rigid motion (\eg talking, smiling, etc.), and sensor noise, since they may preside outside the predefined bandlimits or with uniform power spectra.
Even if noise exhibits periodic tendencies within the bandlimits for some samples, those features would produce poor signals on other samples.
Therefore, end-to-end unsupervised approaches are particularly well-suited for periodic problems.

\section{Conclusions}
We proposed a novel non-contrastive learning approach for end-to-end unsupervised signal regression, with specific experiments on blood volume pulse estimation from face videos.
This SiNC framework effectively learns powerful visual features with only loose frequency constraints.
We demonstrated this by training accurate rPPG models using non-rPPG data and our simple loss functions.
Given the subtlety of the rPPG signal, we believe our work can be extended to other signal regression tasks in the domain of remote vitals estimation.
\section*{Acknowledgements}
This research was sponsored by the Securiport Global Innovation Cell, a division of Securiport LLC. Commercial equipment is identified in this work in order to adequately specify or describe the subject matter. In no case does such identification imply recommendation or endorsement by Securiport LLC, nor does it imply that the equipment identified is necessarily the best available for this purpose. The opinions, findings, and conclusions or recommendations expressed in this publication are those of the authors and do not necessarily reflect the views of our sponsors.

{\small
\bibliographystyle{ieee_fullname}
\bibliography{biblio}
}
\appendix

\section{Appendix}

\subsection{rPPG Signal Quality on CelebV-HQ}\label{sec:app_celebv}
\begin{figure}
    \centering
    \includegraphics[width=1\linewidth]{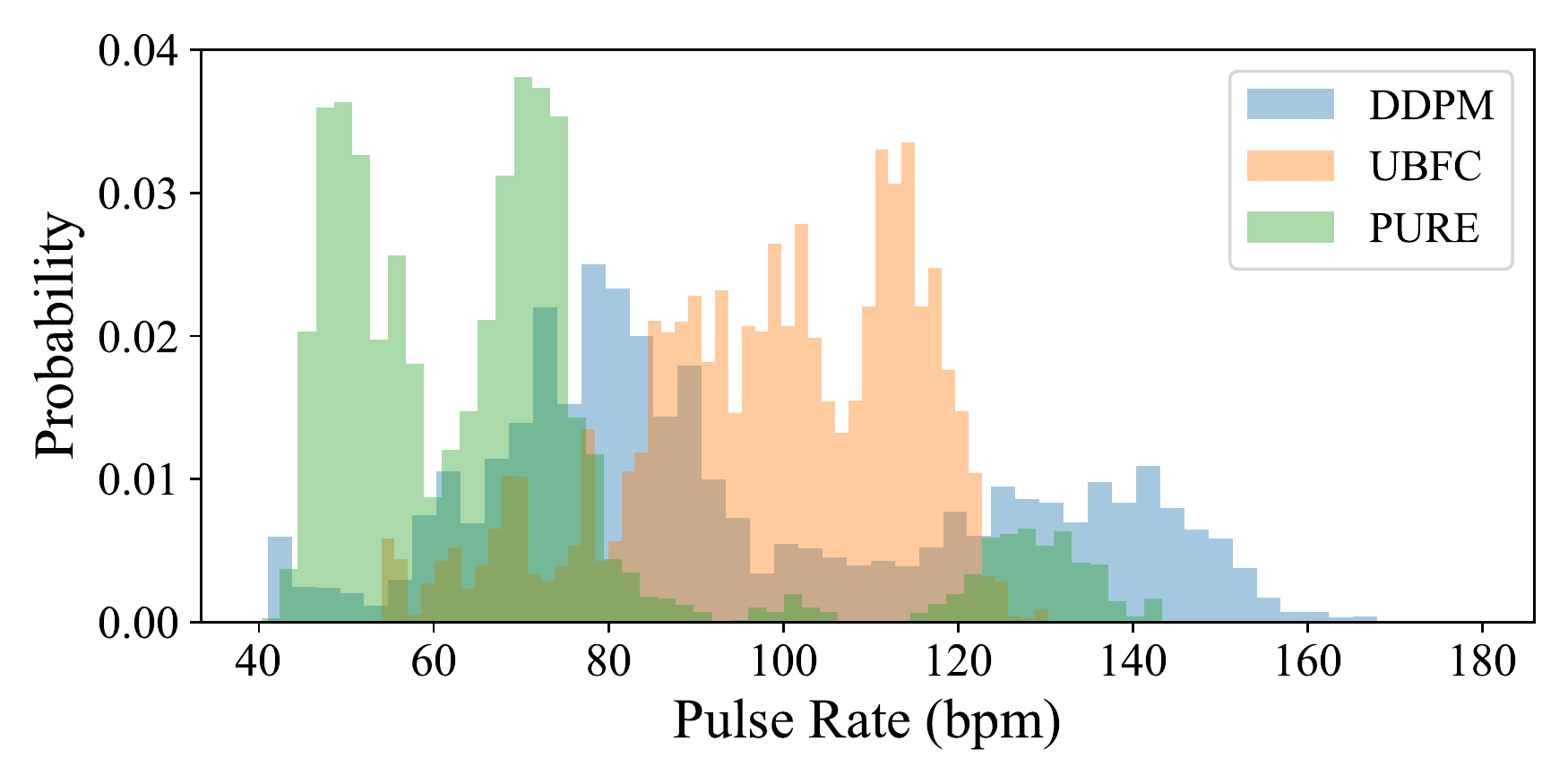}
    \caption{Distribution of ground truth pulse rates over the three main rPPG datasets explored in this paper.}
    \label{fig:app_dataset_HRs}
\end{figure}

Given the enormous quantity of video data present in CelebV-HQ, we were surprised to find that SiNC was unable to correctly learn the pulse signal. To better understand why training on CelebV-HQ is challenging, we predicted rPPG signals with the POS~\cite{Wang2017} algorithm, which is a reputable baseline approach that tends to transfer well across different sources of video data.
We calculated the signal-to-noise ratio (SNR) from all of the predictions and compared them with predictions on traditional rPPG datasets. Figure \ref{fig:app_dataset_HRs} shows the histograms of SNRs for each dataset. The SNR for CelebV-HQ is much lower than the other datasets, indicating a lower signal quality. The drop in quality is likely due to video compression, which may have even occurred multiple times before download.

\begin{figure}
    \centering
    \includegraphics[width=1\linewidth]{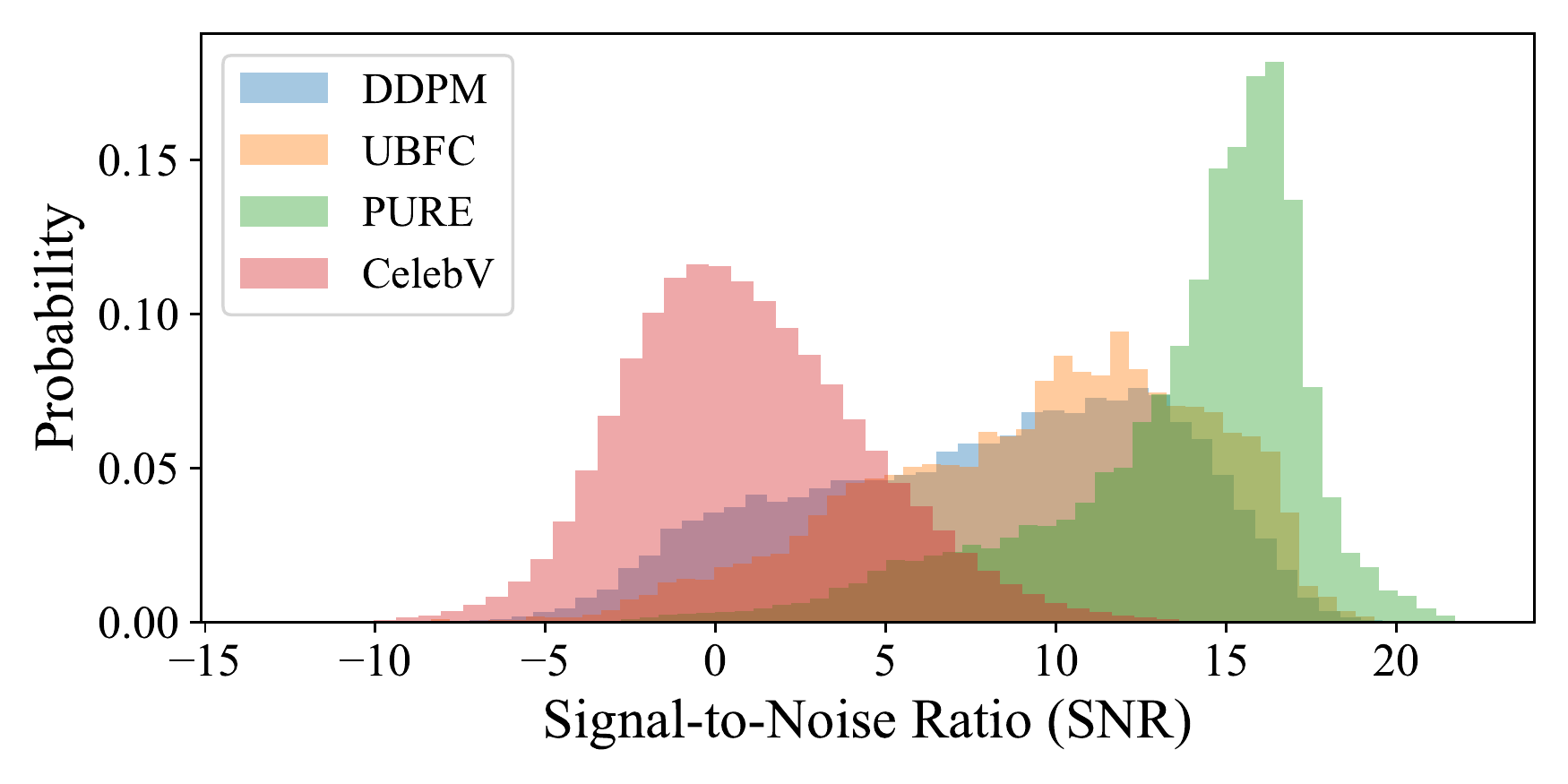}
    \caption{Distribution of SNR values from the POS algorithm over the three main rPPG datasets and CelebV. Although the downloaded videos are of the highest quality available, there is a clear downwards shift in signal quality due to compression and diverse settings. See how this impacted large-scale unsupervised learning with SiNC in Sec. \ref{sec:celebv_training}.}
    \label{fig:app_celebv}
\end{figure}

\subsection{Justification for Frequency Bounds}
To justify our selection of the lower and upper bounds of 40 bpm and 180 bpm, we plotted the distribution of ground truth pulse rates over DDPM, UBFC, and PURE. In general we see that very few pulse rates approach 40, and the highest pulse rates are just beyond 160 bpm.

\subsection{Allowing Second Harmonic in Sparsity Loss}
Many rPPG papers allow signal power in the second harmonic when evaluating their approaches~\cite{DeHaan2013,Nowara_BOE_2021}. We chose not to incorporate higher harmonics to keep the sparsity loss simple and avoid the risk of amplifying the dicrotic notch of the waveform. We verified this empirically by training and testing models on PURE while allowing energy in the second harmonic. The performance dropped due to the peak frequency occasionally occurring in the second harmonic (MAE of $3.29 \pm 1.69$). For the purpose of pulse rate estimation, including the dicrotic notch can actually introduce inaccuracies.

\subsection{Impact of Batch Size}\label{sec:app_batch_size}
The variance component of the loss depends on the batch size, since a normalized sum over the batch is calculated. To verify that the batch size can safely be reduced for memory-constrained environments we performed an ablation study with batch sizes in \{5, 10, 15, 20\} when training on the PURE dataset. For within-dataset testing, models gave MAEs of \{$0.73 \pm 0.08$, $0.65 \pm 0.09$, $1.75 \pm 1.39$, $0.61 \pm 0.06$\}, respectively. Overall, the batch size does not seem to have a large influence on performance.

\subsection{Augmentations are Critical}\label{sec:app_no_augmentations}
Several augmentations are used while training SiNC, some of which can even influence the underlying pulse rate distributions (see frequency augmentations in Sec. \ref{sec:augmentations}). To verify that the augmentations play a key role, we trained models without them on PURE. Models trained without augmentations gave a MAE of $33.86 \pm 0.83$. Therefore, training with augmentations is critical within the SiNC approach for models to converge to the blood volume pulse.
\end{document}